\documentclass{article}

\usepackage{arxiv}
\usepackage{hyperref}
\usepackage{amsfonts}       % blackboard math symbols
\usepackage[numbers]{natbib}
\def\bng{\bngx}

\font\bngx=bang10

\def\*#1*#2{o\null{#2}{#1}}

\def\sh#1{\setbox0=\hbox{#1}%
     \kern-.02em\copy0\kern-\wd0
     \kern.04em\copy0\kern-\wd0
     \kern-.02em\raise.0433em\box0 }
\usepackage[ruled,vlined,linesnumbered]{algorithm2e}
\usepackage{amsmath}
\usepackage{graphicx}

\title{A Novel Word Pair-based Gaussian Sentence Similarity Algorithm For Bengali Extractive Text Summarization}

\author{
 Fahim Morshed \\
  Institute of Information Technology\\
  University of Dhaka\\
  Dhaka, Bangladesh \\
  \texttt{f.morshed.opee@gmail.com} \\
   \And
 Md. Abdur Rahman \\
  Centre for Advanced Research in Science\\
  University of Dhaka\\
  Dhaka, Bangladesh \\
  \texttt{mukul.arahman@gmail.com} \\
  \And
 Sumon Ahmed\thanks{corresponding author} \\
  Institute of Information Technology\\
  University of Dhaka\\
  Dhaka, Bangladesh \\
  \texttt{sumon@du.ac.bd} \\
}

\begin{document}
\maketitle
\begin{abstract}
Extractive Text Summarization is the process of selecting the most representative parts of a larger text without losing any key information. Recent attempts at extractive text summarization in Bengali, either relied on statistical techniques like TF-IDF or used naive sentence similarity measures like the word averaging technique. All of these strategies suffer from expressing semantic relationships correctly. Here, we propose a novel Word pair-based Gaussian Sentence Similarity (WGSS) algorithm for calculating the semantic relation between two sentences. WGSS takes the geometric means of individual Gaussian similarity values of word embedding vectors to get the semantic relationship between sentences. It compares two sentences on a word-to-word basis which rectifies the sentence representation problem faced by the word averaging method. The summarization process extracts key sentences by grouping semantically similar sentences into clusters using the Spectral Clustering algorithm. After clustering, we use TF-IDF ranking to pick the best sentence from each cluster. The proposed method is validated using four different datasets, and it outperformed other recent models by 43.2\% on average ROUGE scores (ranging from 2.5\% to 95.4\%). It is also experimented on other low-resource languages i.e. Turkish, Marathi, and Hindi language, where we find that the proposed method performs as similar as Bengali for these languages. In addition, a new high-quality Bengali dataset is curated which contains 250 articles and a pair of summaries for each of them. We believe this research is a crucial addition to Bengali Natural Language Processing (NLP) research and it can easily be extended into other low-resource languages. We made the implementation of the proposed model and data public on \href{https://github.com/FMOpee/WGSS}{https://github.com/FMOpee/WGSS}.
\end{abstract}

% keywords can be removed

\section{Introduction}\label{sec:introduction}
Text Summarization is the process of shortening a larger text without losing any key information to increase readability and save time for the readers. However, manually summarizing very large texts is a counter-productive task due to it being more time-consuming and tedious. So, developing an Automatic Text Summarization (ATS) method that can summarize larger texts reliably is necessary to alleviate this manual labour \citep{Widyassari-2022-rev-ats-tech-met}. Using ATS to summarize textual data is thus becoming very important in various fields such as news articles, legal documents, health reports, research papers, social media content etc. ATS helps the readers to quickly and efficiently get the essential information without needing to read through large amounts of texts \citep{wafaa-2021-summary-comprehensive-review}. ATS is being utilized in various fields, from automatic news summarization, content filtering, and recommendation systems to assisting legal professionals and researchers in going through long documents by condensing vast amounts of information. It plays a critical role in personal assistants and chatbots, providing condensed information to users quickly and efficiently \citep{tas-2017-rev-text-sum-2}.

ATS techniques can be broadly classified into extractive and abstractive strategies \citep{tas-2017-rev-text-sum-2}. Extractive summarization, which is the focus of this paper, works by selecting a subset of sentences or clauses from the source document maintaining the original wording and sentence structure \citep{moratanch-2017-extractive-review}. In contrast, abstractive summarization involves synthesising new texts that reflect information from the input document but do not copy from it, similar to how a human summarizes a text document \citep{Moratanch-2016-abstractive-rev}. Both of the methods have their relative advantages and disadvantages. The abstractive summarization can simulate the human language pattern very well thus increasing the natural flow and readability of the summary. However, it requires a vast amount of training data and computational resources to be accurate. On the other hand, the extractive method requires much less computation than the abstractive method while also containing more key information from the input \citep{gupta-2010-extractive-rev} but lacking in natural flow. It is difficult to find quality training data in Bengali due to it being a low-resource language, therefore we attempted to implement an extractive strategy. At the same time, we also provided a quality Bengali summarization dataset.

The key approach to extractive summarization is implementing a sentence selection method to classify which sentences will belong in the summary. For this purpose, various ranking-based methods were used to rank the sentences and identify the best sentences as the summary. These ranking methods used indexing \citep{Baxendale_1958_firstsummarization}, statistical \citep{edmundson_1969_earlysum} or Term Frequency-Inverse Document Frequency (TF-IDF) \citep{das-2022-tfidf,sarkar-2012-tfidf-2} based techniques to score the sentences and select the best scoring ones. Early attempts at Bengali text summarization also relied on TF-IDF scoring to select the best scoring sentences to form the summary \citep{Akter-2017-tfidf-3, sarkar-2012-tfidf-2}. These approaches, while simple, faced challenges in capturing the true meaning of sentences. This is because TF-IDF-based methods treat words as isolated terms resulting in synonyms of words being regarded as different terms \citep{tas-2017-rev-text-sum-2}. To capture the semantic relationships between sentences, graph-based extractive methods are effective due to the use of sentence similarity graphs in their workflow \citep{wafaa-2021-summary-comprehensive-review}. Graph-based methods represent the sentences as nodes of a graph, and the semantic similarity between two sentences as the edge between the nodes \citep{moratanch-2017-extractive-review}. Popular graph-based algorithms like LexRank \citep{Erkan-lexRank-2004} and TextRank \citep{mihalcea-2004-textrank} build graphs based on the cosine similarity of the bag-of-word vectors. LexRank uses PageRank \citep{page-PageRank-1999} method to score the sentences from the graph while TextRank uses random walk to determine which sentences are the most important to be in the summary. Graph-based methods like TextRank and LexRank offer a robust way to capture sentence importance and relationship, ensuring that the extracted summary covers the key information while minimizing redundancy \citep{wafaa-2021-summary-comprehensive-review}.

Clustering-based approaches are a subset of graph-based approaches to extractive text summarization. Here, sentences are grouped into clusters based on their semantic similarity to divide the document into topics, and one representative sentence from each cluster is chosen to form the summary \citep{Mohan-2022-topic-modeling-rev-clustering}. Clustering reduces redundancy by ensuring that similar sentences are grouped together and only the most representative sentence is selected. This method is effective in the summarization of documents with multiple topics or subtopics by picking sentences from each topic \citep{mishra-2014-clustering-topiuc-coverage}. COSUM \citep{alguliyev-2019-cosum} is an example of this method where the summarization is achieved using k-means clustering on the sentences and picking the most salient sentence from each cluster to compile in the final summary.

Despite the advancements of ATS in other languages, it remains an under-researched topic for Bengali due to Bengali being a low-resource language. Graph-based methods were introduced in Bengali to improve summarization quality by incorporating sentence similarity but they were still limited by the quality of word embeddings used for the Bengali language. With the advent of word embedding models like FastText \citep{grave-etal-2018-fasttext}, it became possible to represent words in a vector space model, thus enabling more accurate sentence similarity calculations. However, existing models that use word embeddings, such as the Sentence Average Similarity-based Spectral Clustering (SASbSC) method \citep{roychowdhury-etal-2022-spectral-base}, encountered issues with sentence-similarity calculation when averaging word vectors to represent the meaning of a sentence with a vector. This method failed in most similarity calculation cases because words in a sentence are complementary to each other rather than being similar, leading to inaccurate sentence representations after averaging these different word vectors. As a result, word-to-word relationships between sentences get lost, reducing the effectiveness of the method.

In this paper, we propose a new clustering-based text summarization approach to address the challenge of calculating sentence similarity accurately. Our method improves upon previous attempts at graph-based summarization methods \citep{chowdhury-etal-2021-tfidf-clustering, roychowdhury-etal-2022-spectral-base} by focusing on the individual similarity between word pairs in sentences rather than averaging word vectors. We showed that the use of individual similarity between word pairs greatly improves the accuracy, coverage and reliability of the output summaries due to having a deeper understanding of the semantic similarity between sentences. To calculate sentence similarity, we used the geometric mean of individual word similarities. The individual word similarities were achieved using the Gaussian kernel function on a pair of corresponding word vectors from each sentence. The word pairs are selected by finding the word vector with the smallest Euclidean distance from the target sentence. Thus, we get the semantic similarity between two sentences which is used to build an affinity matrix to graphically represent the relationship between the sentences. This graph is clustered into groups to divide the document into distinct topics. One sentence from every cluster is selected to reduce redundancy and increase topic coverage. This method consistently outperforms other graph-based text summarization methods such as BenSumm \citep{chowdhury-etal-2021-tfidf-clustering}, LexRank \citep{Erkan-lexRank-2004}, SASbSC \citep{roychowdhury-etal-2022-spectral-base} using four datasets on ROUGE metrics \citep{lin-2004-rouge} as shown in Fig. \ref{fig:radarchart} and Table \ref{tab:result_comparison-1}. This method performs well in other low-resource languages also such as Hindi, Marathi and Turkish due to the language-independent nature, as shown in Table \ref{tab:other_language}.

The main contributions of this paper are:
(I) Proposed a new way to calculate the similarity between two sentences.
(II) Contributes a novel methodology for extractive text summarization for the Bengali language; by improving sentence similarity calculations and enhancing clustering techniques.
(III) It offers a generalizable solution for creating less redundant and information-rich summaries across languages.
(IV) It provides a publicly available high-quality dataset of 500 human-written summaries.

The rest of the paper is organized as follows: The Related works and Methodology are described in section \ref{sec:literature-review} and \ref{sec:methodology} respectively. Section \ref{sec:result} illustrates the result of the performance evaluation for this work. Section \ref{sec:discussion} discusses the findings of the paper in more depth, and section \ref{sec:conclusion} concludes the paper.
\section{Related Work}\label{sec:literature-review}
Text summarization has been an important necessity for textual data consumption for a long time because of its ability to compress a given input text into a shortened version without losing any key information. For this reason, automating the text summarization process has been a research problem for NLP. Thus researchers attempted automatic text summarization for a long time. At first, indexing-based text summarization methods were attempted such as the work by \cite{Baxendale_1958_firstsummarization}. This method scored sentences based on the presence of indexing terms in the sentence and picked the sentences with the best score. However, this method failed to capture the topic and essence of the input text as we wouldn't have the topic keywords of an unforeseen input document. To solve this issue, text summarization with statistical methods like TF-IDF became very popular due to its ability to capture the important topic words of a document in an unsupervised manner. \cite{edmundson_1969_earlysum} proposed a method which can focus on the central topic of a document by using the TF-IDF measure. It uses two metrics, Term Frequency (how many times a term appears in the input) and Inverse Document Frequency (inverse of how many documents the term appears in a large text corpus) to calculate the importance of a term in a document. Using TF-IDF helps to identify the words that are common in the input text but not as common in the language in general thus classifying them as the central topic of the document. But this method is also error-prone due to its consideration of every word as a unique isolated term without any semantic relation with other words. This leads to the method often missing some central topic if it gets divided into too many synonyms.

The problems faced by topic-based summarization methods were alleviated by graph-based extractive text summarization methods which brought new modern breakthroughs. Graph-based methods like LexRank \cite{Erkan-lexRank-2004} and TextRank \cite{mihalcea-2004-textrank} were able to capture the relationship between sentences more accurately due to the use of sentence similarity graphs in their process. LexRank \cite{Erkan-lexRank-2004} calculates the similarity graph by using the cosine similarity of bag-of-words vectors between two sentences from the input. The most important sentences from the graph are classified using the PageRank \cite{page-PageRank-1999} algorithm on the graph. PageRank ranks those sentences higher and is more similar to other high-ranked sentences. Another graph-based method, TextRank \cite{mihalcea-2004-textrank} also uses a similar approach while building the similarity graph. In the graph for every sentence, TextRank distributed the score of that sentence to its neighbours using a random walk. This process is repeated over and over until the scores of the sentences stop changing. Then the method picks the sentences with the best scores as the summary. Although graph-based methods such as LexRank and TextRank models are ground-breaking compared to their time, they lack a fundamental understanding of the words involved in a sentence due to not using any vector representation of the semantic relationship between the words involved.

To solve the problem of representing semantic relationships, a mathematical abstraction called Word Vector Embedding was conceptualized by the seminal work of \cite{salton-1975-word-vector}. A lexicon, which is essentially the vocabulary of a language or a branch of knowledge, is represented in this context as a word vector space. Word Vector Embedding uses this space as a mathematical abstraction where semantically closer words are positioned nearer to each other in the vector space. The use of word vectors for graph-based text summarization has only recently been attempted \cite{Jain-2017-word-vector-embedding-summary} due to the previous lack of fully developed word embedding datasets in low-resource languages like Bengali.

Although text summarization has been at the forefront of NLP research, text summarization research in Bengali is a more recent development than in other high-resource languages. So, a lot of sophisticated approaches from other languages haven't been attempted in Bengali yet. Earlier Bengali extractive methods have been focused on some derivative of TF-IDF-based text summarization such as the methods developed by \cite{chowdhury-etal-2021-tfidf-clustering}, \cite{das-2022-tfidf}, \cite{sarkar-2012-tfidf} etc. \cite{sarkar-2012-tfidf} used a simple TF-IDF score of each sentence to rank them and pick the best sentences to generate the output summary. \cite{das-2022-tfidf} used weighted TF-IDF along with some other sentence features like sentence position to rank the sentences. \cite{chowdhury-etal-2021-tfidf-clustering} however, used TF-IDF matrix of a document to build a graph and perform hierarchical clustering to group sentences together and pick one sentence from each group. One shortcoming of this method is that the TF-IDF matrix is not semantically equivalent to the actual sentences due to the fundamental issues with TF-IDF. So, the TF-IDF doesn't perfectly represent the semantic relationship between the sentences in the generated graph. Using word vector embedding for Bengali has solved this problem of semantic representation. FastText \cite{grave-etal-2018-fasttext} dataset\footnote{\textit{\href{https://fasttext.cc/docs/en/crawl-vectors.html}{https://fasttext.cc/docs/en/crawl-vectors.html}}} provides good word vector embeddings in 157 languages, including Bengali. Using this dataset, SASbSC \cite{roychowdhury-etal-2022-spectral-base} proposed a model where they replaced all the words from the input with their respective vector, then averaged the word vectors of a sentence to get a vector representation for the sentence. The sentence average vectors are then used to get the similarity between two sentences using the Gaussian similarity function to build an affinity matrix. This affinity matrix is used to cluster the sentences using spectral clustering to group sentences into distinct topics. The summary is generated after picking one sentence from each cluster to reduce redundancy and increase coverage.

\begin{figure}
	\centering
	\includegraphics[width=5in]{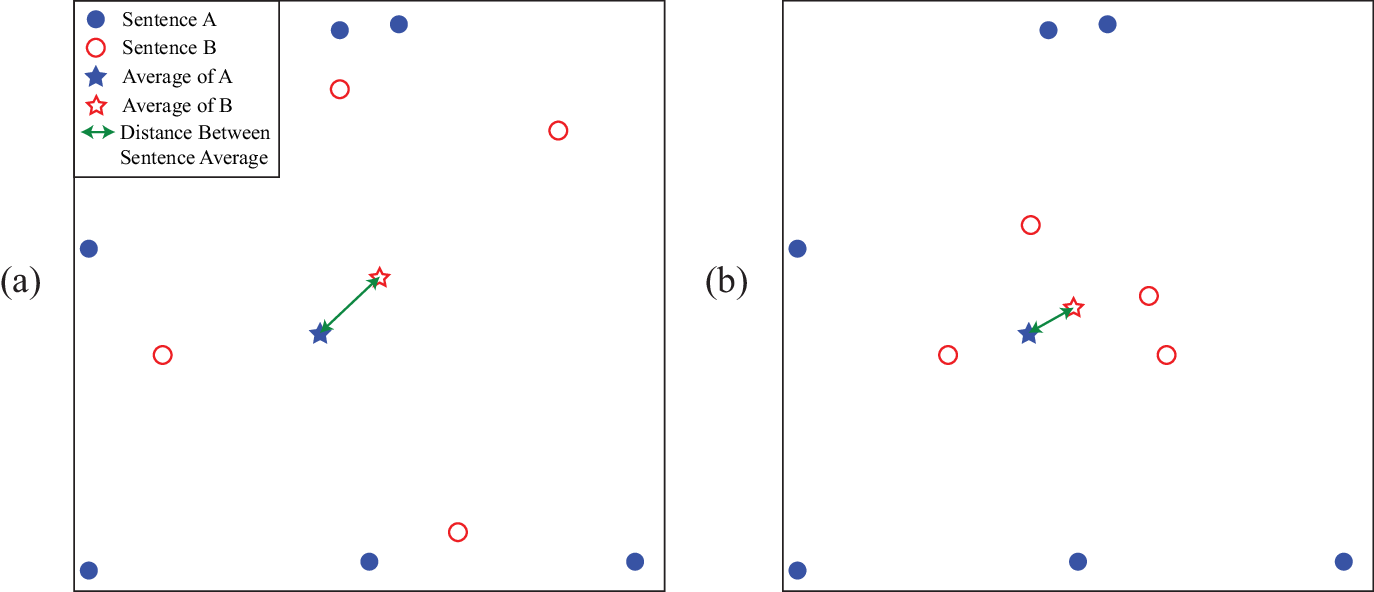}
	\caption{A scenario where the sentence averaging method fails. (a) shows a scenario where the distance between sentence average vectors is larger than (b) despite the word vectors from (a) being more closely related than (b).}
	\label{fig:sarkar-problem}
\end{figure}

The sentence average method suffers critically due to its inability to capture accurate relationships between sentences. This happens due to words in a sentence generally do not have similar meanings to each other, instead, they express different parts of one whole meaning of a sentence. This makes the words more complementary instead of being similar leading to word vectors being scattered throughout the word vector space. This characteristics makes the sentence average vectors always tending towards the centre and not representing the semantic similarity accurately. An example of this happening is shown in Fig. \ref{fig:sarkar-problem} where the distance between the sentence average vectors is misleading. In the figure, scenario (a) shows two sentences where word vectors are very closely corresponding with each other. On the other hand, scenario (b) shows two sentences without any significant word correspondence. But scenario (a) has a larger distance between sentence average vectors than scenario (b) despite having more word correspondence. This larger distance makes the Gaussian similarity lower between the sentences due to the inverse exponential nature of the function. The lower similarity leads to the graphical representation being less accurate and thus failing to capture the true semantic relationship within the sentences. This shortcoming of the method has been one of the key motivations for this research.
\section{Methodology}\label{sec:methodology}
\begin{figure}
	\centering
	\includegraphics[width=5in]{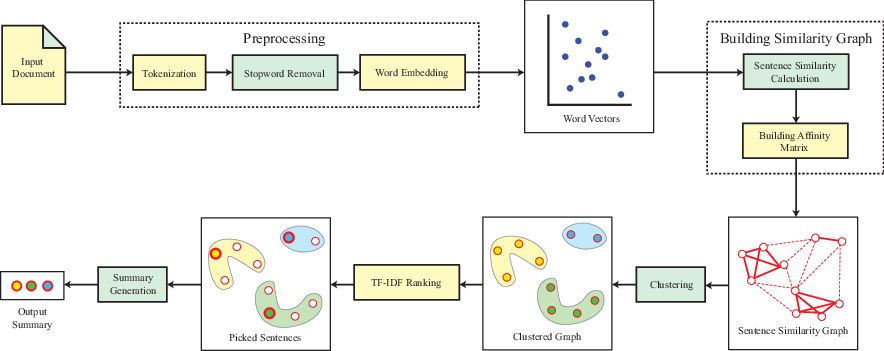}
	\caption{Process Flow Diagram}
	\label{fig:process-flow-diagram}
\end{figure}

The summarization process followed here is comprised of two basic steps, grouping all the close sentences based on their meaning to minimize redundancy, and then picking one sentence from each group to maximize sentence coverage. To group semantically similar sentences into clusters, we build a sentence similarity graph and perform spectral clustering on it \cite{roychowdhury-etal-2022-spectral-base}. The sentence similarity graph is produced using a novel sentence similarity calculation algorithm (see Algorithm \ref{alg:similarity}) that uses the geometric mean of Gaussian similarity between individual word pairs from the two sentences. The Gaussian similarity is calculated using the vector embedding representations of the words. Secondly, we used TF-IDF scores to pick the highest-ranked sentences as in \cite{Akter-2017-tfidf-3, das-2022-tfidf, sarkar-2012-tfidf, sarkar-2012-tfidf-2} from each cluster (see Algorithm \ref{alg:summary}). Therefore The summarization process followed here involves Pre-processing, Sentence similarity calculation, Clustering and Summary generation. These steps are illustrated in Fig. \ref{fig:process-flow-diagram} and further discussed in the following subsections. 
\subsection{Preprocessing}\label{subsec:preprocessing}
Preprocessing is the standard process of NLP that transforms raw human language inputs into a format that can be used by a computer algorithm. In this step, the document is transformed into a few sets of vectors where each word is represented with a vector, each sentence is represented as a set of vectors and the whole document as a list containing those sets. To achieve this representation, the preprocessing follows three steps. These are tokenization, stop word removal, and word embedding. 

Tokenization is the step of dividing an input document into sentences and words to transform it into a more usable format. Here, the input document is represented as a list of sentences and the sentences are represented as a list of words. Stop words, such as prepositions and conjunctions, add sentence fluidity but don’t carry significant meaning. Removing these words allows the algorithm to focus on more impactful words. Word Embedding is the process of representing words as vectors in a vector space. In this vector space, semantically similar words are placed closer together so that the similarity relation between words is expressed in an abstract mathematical way. Each word from the tokenized and filtered list is replaced with its corresponding vectors. Following these steps, the input document is transformed into a set of vectors to be used in sentence similarity calculation. The word embedding dataset \cite{grave-etal-2018-fasttext} can provide word vectors for all the possible variations of a root word instead of just the root words. So stemmed root words would result in loss of information from the input document. To avoid this, no word stemming was performed on the tokenized and filtered document. Some examples of words present in the embedding dataset are shown in Table \ref{tab:dataset}.
\begin{table}[]
    \centering
	\caption{Examples of word vectors present in FastTexts Dataset}
	\label{tab:dataset}
	\begin{tabular}{|l|l|l|}
		\hline
		\textbf{Language}	&\textbf{Root Word}	&\textbf{Words Present at Dataset} \\ \hline
		English				&Do					&Do, Does, Done, Doing, Undo, Redo etc. \\ \cline{2-3}
							&Brave				&Bravery, Bravest, Bravado, Brave-heart etc. \\ \hline
		Bengali				&{\bng kr}	    	&{\bng kreb, kerech, kir, krb, kr/ta, ikRya} etc. \\ \cline{2-3}
							&{\bng man}			&{\bng pRman, Apman, sm/man, mHamanY, manniiy} etc. \\ \hline 
	\end{tabular}
\end{table}
\subsection{Sentence Similarity Calculation}\label{subsec:sentence-similarity-calculation}
Sentence similarity is the key criterion to build a graphical representation of the sentences in terms of their semantic relationships. This graphical representation is expressed via an affinity matrix to cluster the semantically similar sentences. The rows and columns in the affinity matrix represent the sentences and the cells of the matrix represent the similarity between two sentences. For the similarity value, we have to compare two sets of word vectors. The existing methods to compare two sets of vectors, such as Word Averaging Method \cite{roychowdhury-etal-2022-spectral-base}, Earth Movers Distance (EMD) \cite{Rubner-19998-emd}, Procrustes Analysis \cite{Gower-1975-procrustes-distance}, Hausdorff Distance \cite{hausdorff-1914-hausdorff-distance}, all have shortcomings in the context of word vectors. The Word Averaging Method \cite{roychowdhury-etal-2022-spectral-base} fails to represent local word relation within the sentences in most cases (explained in Fig. \ref{fig:sarkar-problem}). EMD \cite{Rubner-19998-emd} attempts to find the lowest "cost" to transform one set of vectors into another. However, it is computationally expensive, especially for high-dimensional word vectors, and the concept of "optimal transport" is not particularly relevant in a linguistic context. Procrustes Analysis \cite{Gower-1975-procrustes-distance} calculates the minimum misalignment after rotating and scaling two vector sets, which is also irrelevant for linguistic applications. The Hausdorff Distance \cite{hausdorff-1914-hausdorff-distance} measures the greatest distance from a point in one set to the closest point in the other set. Although it is less computationally demanding than EMD and Procrustes Analysis, it is easily influenced by outlier words due to its focus on the worst-case scenario. To solve these shortcomings of the mentioned methods, we proposed a novel sentence similarity calculation technique using individual Gaussian similarity of word pairs to construct an affinity matrix. To calculate the sentence similarity between two sentences, we adhere to the following steps.

\begin{figure}
	\centering
	\includegraphics[width=5in]{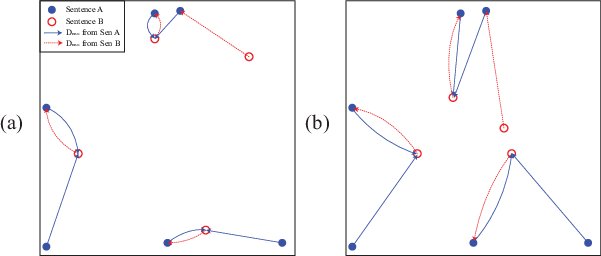}
	\caption{Emphasis of local word correspondence in $D_{msw}$ method. Here, scenario (a) has a larger similarity value due to having a set of smaller $D_{msw}$ than scenario (b)}
	\label{fig:msd}
\end{figure}

Firstly, for every word of a sentence, we find its closest counterpart from the other sentence to build a word pair. The Euclidean distance between the vector representation of the word pairs is defined as the Most Similar Word Distance ($D_{msw}$). The process of calculating the $D_{msw}$ is shown in the Equation \ref{eq:msd}. In this equation, for every word vector $x$, in a sentence $X$, we find the Euclidean distance ( $d(x,y_i)$ ) between the word vectors $x$ and $y_i$ where $y_i$ is a word vector from the sentence $Y$. The lowest possible distance in this set of Euclidean distances is the $D_{msw}$. 
\begin{equation}\label{eq:msd}
	D_{msw}(x,Y) = min(\{d(x,y_i) : y_i \in Y \})
\end{equation}
Then, we calculate the $D_{msw}$ for every word of the two sentences $X$ and $Y$ to make the sentence similarity calculation symmetric. This process is shown in the Equation \ref{eq:mswdset} where $D$ is a set containing all the $D_{msw}$ from both $X$ and $Y$ that would be used in the later steps.
\begin{equation}
	D = \{D_{msw}(x_i,Y) : x_i \in X\} \cup \{D_{msw}(y_j,X) : y_j \in Y\}
	\label{eq:mswdset}
\end{equation}
After this, the word similarity between the word pairs is calculated to get the degree of correspondence between the two sentences. Here, the word similarity is calculated using the Gaussian kernel function for the elements of the set $D$; Gaussian kernel functions provide a smooth, flexible and most representative similarity between two vectors \cite{babud-1986-gaussian}. The process of calculating word similarity ($W_{sim}$) is given in the Equation \ref{eq:wsim}. In this equation, for every element $D_i$ in set $D$, we calculate the Gaussian similarity to obtain word similarity. In the formula for Gaussian similarity, $\sigma$ denotes the standard deviation which is used as a control variable. The standard deviation represents the blurring effect of the kernel function. A lower value for $\sigma$ represents a high noise sensitivity of the function \cite{babud-1986-gaussian}. The value of $\sigma$ was fine-tuned to be $5\times10^{-11}$ which gives the best similarity measurement. The process of fine-tuning is described in the experimentation section (section \ref{subsubsec:sigma}). 
\begin{equation}\label{eq:wsim}
	W_{sim} = \{ exp\left(\frac{-D_i^2}{2\sigma^2}\right) : D_i \in D\}
\end{equation}
Finally, the sentence similarity between the two sentences $Sim(X, Y)$ is calculated using the geometric mean of the word similarities to construct an affinity matrix. The geometric mean makes the similarity value less prone to the effects of outliers thus it makes the calculation more reliable. The geometric mean of each $W_{sim}$ value for the two sentences is simplified in Equation \ref{eq:sent_sim} to make the calculation process more computation friendly. 
\begin{equation}\label{eq:sent_sim}
	\begin{split}
		Sim(X,Y)
		&=  \left(
			\prod_{i=1}^nW_{Sim_i}
		\right)^{1/n}\\
		&=  \left(
			exp\left(\frac{-D_1^2}{2\sigma^2}\right)\cdot
			exp\left(\frac{-D_2^2}{2\sigma^2}\right)\cdot
				\ldots \cdot
			exp\left(\frac{-D_n^2}{2\sigma^2}\right)
		\right)^{1/n}\\
		&=  exp\left(
			-\frac{D_1^2+D_2^2+\ldots+D_n^2}{2n\sigma^2}
			\right)\\
		&=  exp\left(
			-\frac{\sum_{i=1}^nD_i^2}{2n\sigma^2}
		\right)
	\end{split}
\end{equation}
By following the steps described in the above equations, we get a similarity value for two sentences. This value solves the lack of local word correspondence problem faced by the word averaging-based similarity calculation method used in SASbSC \cite{roychowdhury-etal-2022-spectral-base}. Fig. \ref{fig:msd} demonstrates the merit of the proposed method. Fig. \ref{fig:msd} uses the same scenario from Fig. \ref{fig:sarkar-problem} and shows that the proposed method can solve the local word correspondence problem faced by the word averaging method. In the figure, the scenario \ref{fig:msd}(a) has a set of smaller $D_{msw}$ than the scenario \ref{fig:msd}(b). Having smaller $D_{msw}$ makes the individual word similarities $W_{sim}$ larger due to the nature of the Gaussian kernel function. These values would result in a higher sentence similarity for the sentences in the scenario \ref{fig:msd}(a) than in the scenario \ref{fig:msd}(b). This representative sentence similarity solves the problem shown in Fig. \ref{fig:sarkar-problem} where the scenario \ref{fig:sarkar-problem}(a) has a larger sentence average distance than \ref{fig:sarkar-problem}(b) resulting in \ref{fig:sarkar-problem}(a) having a smaller sentence similarity than \ref{fig:sarkar-problem}(b).

The whole process of sentence similarity calculation is shown in the Algorithm \ref{alg:similarity} where we calculate an affinity matrix using the input word vector list. We took the most similar word distance ($D_{msw}$) (lines 5--10 and 15--20) for each word (lines 4 and 14) of a sentence pair (line 1). The sum of $D^2$ from Equation \ref{eq:sent_sim} is calculated (lines 11 and 21) to be used in the calculation of sentence similarity (line 24). 
\begin{algorithm}[]
	\caption{Sentence Similarity}
	\label{alg:similarity}
	\SetAlgoLined
	
	\SetKwFunction{FnSim}{similarity}
	\SetKwProg{Fn}{Function}{:}{end}
	\Fn{\FnSim{sentence$_i$,sentence$_j$}}{
		$D_{\text{Square}} \gets 0$\;
		$n \gets 0$\;
		\ForEach{word$_i$ \textbf{in} sentence$_i$}{
			$D_{\text{msw}} \gets \infty$\;
			\ForEach{word$_j$ \textbf{in} sentence$_j$}{
				\If{\texttt{Distance}(word$_i$, word$_j$) $< D_{\text{msw}}$}{
					$D_{\text{msw}} \gets$ \texttt{Distance}(word$_i$, word$_j$)\;}}
			$D_{\text{Square}} \gets D_{\text{Square}} + D_{\text{msw}}^2$\;
			$n \gets n + 1$\;}
		\ForEach{word$_j$ \textbf{in} sentence$_j$}{
			$D_{\text{msw}} \gets \infty$\;
			\ForEach{word$_i$ \textbf{in} sentence$_i$}{
				\If{\texttt{Distance}(word$_i$, word$_j$) $< D_{\text{msw}}$}{
					$D_{\text{msw}} \gets$ \texttt{Distance}(word$_i$, word$_j$)\;}}
			$D_{\text{Square}} \gets D_{\text{Square}} + D_{\text{msw}}^2$\;
			$n \gets n + 1$\;}
		similarity $\gets \exp \left( \frac{- D_{\text{Square}}}{2n\sigma^2} \right)$\;
		\Return similarity;}
\end{algorithm}
\subsection{Clustering}\label{subsec:clustering}
Clustering is a key cornerstone of the proposed method where we aim to cluster semantically similar sentences together to divide the document into multiple sub-topics. Clustering the document minimizes redundancy in the output summary by ignoring multiple sentences from the same sub-topic. For clustering, spectral and DBSCAN methods are used frequently due to their capability to identify irregular cluster shapes. However, in the case of smaller input documents, DBSCAN would fail due to the low density of the sentences. So spectral clustering was found to perform better than DBSCAN in summarization tasks \cite{roychowdhury-etal-2022-spectral-base}.

On the contrary, spectral clustering takes the affinity matrix of a graph as input and returns the grouping of graph nodes by transforming the graph into its eigenspace \cite{vonLuxburg-2007-spectral-tutorial}. The Equation \ref{eq:affinity} shows the process of building an affinity matrix. Here, for every sentence pair $S_i$ and $S_j$, we calculate their sentence similarity and place it in both $A_{ij}$ and $A_{ji}$ of the affinity matrix $A$.
\begin{equation}\label{eq:affinity}
	A_{ij}=A_{ji}=Sim(S_i,S_j)
\end{equation}
The affinity matrix is clustered into $k=\lceil N \times p \rceil$ groups to achieve an output summary. Here, $N$ is the number of sentences in the input document and $p$ is the proportion of the expected summary to the input. The value of $p$ can be set depending on the size of the input document. 
\subsection{Summary Generation}\label{subsec:summary-generation}
Output summary is generated by selecting one sentence from each cluster to minimize topic redundancy and maximize topic coverage. To select one sentence from a cluster, we perform TF-IDF ranking on the sentences inside a cluster and pick the sentence with the highest TF-IDF score. The TF-IDF value for a word is achieved by multiplying how many times the word appeared in the input document (Term Frequency, TF) and the inverse of how many documents the word appears in a corpus (Inverse Document Frequency, IDF). We take the sum of the TF-IDF values for all words in a sentence to get the score for that sentence. The process of scoring sentences is shown in the Equation \ref{eq:tfidf} where, for each word $W_i$ in a sentence $S$ and a corpus $C$, we calculate the TF-IDF score of a sentence.
\begin{equation}\label{eq:tfidf}
	\text{TFIDF}(S) = \sum_{i=1}^{\text{length}(S)}\text{TF}(W_i) \times \text{IDF}(W_i,C)	
\end{equation}
The sentences with the best TF-IDF score from each cluster are then compiled as the output summary in their order of appearance in the input document to preserve the original flow of information. The process of generating output summary is further expanded in the Algorithm \ref{alg:summary}. Where after the clustering step (line 2), we took the TF-IDF score (line 7) of each sentence in a cluster (line 6). For each cluster (line 4), we pick the best-scoring sentence (line 9). These sentences are then ordered (line 11) and concatenated (lines 13--15) to generate the output summary.
\begin{algorithm}[]
	\caption{Summary Generation}
	\label{alg:summary}
	\SetAlgoLined
	$k \gets \lceil$ length($A$) $* p \rceil$  \hspace{2cm} \tcc*[h]{$0<p<1$}\;
	clusters $\gets$ \texttt{spectral\_clustering(adjacency = $A$, $k$)}\;
	indexes $\gets \{\}$\;
	\ForEach{cluster$_i$ \textbf{in} clusters}{
		TFIDF $\gets \{\}$\;
		\ForEach{index \textbf{in} cluster$_i$}{
			TFIDF.append(\texttt{tfidf\_sum(sentences[index])})\;
		}
		indexes.append(\texttt{indexof(max(TFIDF))})\;
	}
	\texttt{sort(indexes)}\;
	$S \gets $""\;
	\ForEach{$i$ \textbf{in} indexes}{
		$S \gets S +$ sentences[$i$]\;
	}
	\Return $S$\;
\end{algorithm}
\section{Result}\label{sec:result}
The performance of the proposed method has been compared against three Bengali text summarization methods to evaluate the correctness of generated summaries. The benchmark methods, are BenSumm \cite{chowdhury-etal-2021-tfidf-clustering}, LexRank \cite{Erkan-lexRank-2004} and SASbSC \cite{roychowdhury-etal-2022-spectral-base}. All of these methods have been evaluated using four datasets to test the robustness of the model for Bengali text summarization. For evaluation, the Recall-Oriented Understudy for Gisting Evaluation (ROUGE) \cite{lin-2004-rouge} metric has been used. Details about the models, datasets and evaluation metrics are provided in the following sub-sections.
\subsection{Text Summarization Models}\label{subsec:text-summarization-models}
We implemented Bensumm \cite{chowdhury-etal-2021-tfidf-clustering} and SASbSC \cite{roychowdhury-etal-2022-spectral-base}, two recent Bengali extractive models, and LexRank \cite{Erkan-lexRank-2004}, a popular benchmarking model for extractive text summarization to evaluate the effectiveness of the proposed WGSS method. These methods are further discussed in the following section.

\textbf{WGSS} is the proposed model for this research. We find the Gaussian similarity for word pairs from two sentences and take their geometric mean to get the similarity between two sentences. We use the similarity value to perform spectral clustering to group sentences and extract representative sentences using the TF-IDF score. The extracted sentences are used to generate the output summary which minimizes redundancy and maximizes coverage.

\textbf{SASbSC} \cite{roychowdhury-etal-2022-spectral-base} is the first method that introduced the approach of clustering sentences using sentence similarity. However, it uses the average of word vectors in a sentence for calculating similarity. After clustering the sentences based on their similarity, cosine similarity between the average vectors is used to pick the best sentence from a cluster.

\textbf{BenSumm} \cite{chowdhury-etal-2021-tfidf-clustering} is another recent research that describes an extractive and an abstractive text summarization techniques. We compared the extractive technique with our model to ensure a fair and balanced comparison. Here, a similarity matrix is built using TF-IDF which groups the sentences using agglomerative clustering. A Github implementation\footnote{\textit{\href{https://github.com/tafseer-nayeem/BengaliSummarization}{https://github.com/tafseer-nayeem/BengaliSummarization}}} provided by the authors is used in the comparison process.

\textbf{LexRank} \cite{Erkan-lexRank-2004} uses a TF-IDF based matrix and Googles PageRank algorithm \cite{page-PageRank-1999} to rank sentences. Then the top-ranked sentences are selected and arranged into summary. An implemented version of this method is available as lexrank\footnote{\textit{\href{https://pypi.org/project/lexrank/}{https://pypi.org/project/lexrank/}}} which is used in the comparison process using a large Bengali Wikipedia corpus\footnote{\textit{\href{https://www.kaggle.com/datasets/shazol/bangla-wikipedia-corpus}{https://www.kaggle.com/datasets/shazol/bangla-wikipedia-corpus}}}.
\subsection{Evaluation Datasets}\label{subsec:evaluation-datasets}
We used four evaluation datasets with varying quality, size and source to examine the robustness of the methods being tested. The first dataset is a \textbf{self-curated} extractive dataset that we developed to evaluate the performance of our proposed method. An expert linguistic team of ten members summarized 250 news articles of varying sizes to diversify the dataset. Each article is summarized twice by two different experts to minimize human bias in the summary. In total, 500 different document-summary pairs are present in this dataset. The dataset is publicly available on Github\footnote{\textit{\href{https://www.github.com/FMOpee/WGSS/}{https://www.github.com/FMOpee/WGSS/}}} for reproducibility.

The second dataset is collected from \textbf{Kaggle}\footnote{\textit{\href{https://www.kaggle.com/datasets/towhidahmedfoysal/bangla-summarization-datasetprothom-alo}{https://www.kaggle.com/datasets/towhidahmedfoysal/bangla-summarization-datasetprothom-alo}}} which is a collection of summary-article pair from ``The Daily Prothom Alo" newspaper. The dataset is vast in size however the quality of the summaries is poor. For our experimentations, the articles smaller than 50 characters are discarded from the dataset. The articles with unrelated summaries are also removed from the dataset to improve its quality. After filtering, a total of 10,204 articles remained, each with two summaries in the dataset.

The third dataset we used for evaluation is \textbf{BNLPC} which is a collection of news article summaries \cite{Hque-2015-BNLPC-Dataset}. This was collected from GitHub\footnote{\textit{\href{https://github.com/tafseer-nayeem/BengaliSummarization/tree/main/Dataset/BNLPC/Dataset2}{https://github.com/tafseer-nayeem/BengaliSummarization/tree/main/Dataset/BNLPC/Dataset2}}} for experimentation that contains one hundred articles with three different summaries for each article.

The fourth dataset is collected from \textbf{Github}\footnote{\textit{\href{https://github.com/Abid-Mahadi/Bangla-Text-summarization-Dataset}{https://github.com/Abid-Mahadi/Bangla-Text-summarization-Dataset}}}. The dataset contains 200 documents each with two human-written summaries. These documents were collected from several different Bengali news portals. The summaries were generated by linguistic experts which ensures the high quality of the dataset. 
\subsection{Evaluation Metrics}\label{subsec:evaluation-metrics}
To evaluate the correctness of generated summaries against human written summaries, ROUGE metric \cite{lin-2004-rouge} is used. The method compares a reference summary with a machine-generated summary to evaluate the alignment between the two. It uses N-gram-based overlapping to evaluate the quality of generated summaries. The Rouge package\footnote{\textit{\href{https://pypi.org/project/rouge/}{https://pypi.org/project/rouge/}}} is used to evaluate the proposed models against human-written summaries. The package has three different metrics for comparison of summaries. These are are:
\begin{enumerate}
	\item \textbf{ROUGE-1} uses unigram matching to find how similar two summaries are. It calculates the total common characters between the summaries to evaluate the performance. But using this metric also can be misleading as very large texts will share a very high proportion of uni-grams between them.
	\item \textbf{ROUGE-2} uses bi-gram matching to find how similar the two summaries are in a word level. Having more common bigrams between two summaries indicates a deeper syntactic similarity between them. Using this in combination with the ROUGE-1 is a standard practice to evaluate machine-generated summaries \cite{wafaa-2021-summary-comprehensive-review}.
	\item \textbf{ROUGE-LCS} finds the longest common sub-sequence between two summaries to calculate the rouge scores. It focuses on finding similarities in the flow of information in the sentence level between two summaries.
\end{enumerate}
\subsection{Experimentation}\label{subsec:experimentation}
For sentence similarity calculation, we experimented using different values for standard deviation ($\sigma$) in Equation \ref{eq:sent_sim} to get the most representative semantic similarity value. We also experimented with sentence extraction methods to pick the most representative sentence from a cluster. For this, lead extraction and TF-IDF ranking strategies were considered. These experimentations are described in the following sections.
\subsubsection{Fine-tuning Standard Deviation ($\sigma$): }\label{subsubsec:sigma}
\begin{figure}[]
	\centering
	\includegraphics[width=5in]{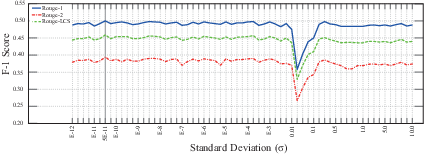}
	\caption{Fine-tuning for different standard deviation ($\sigma$) values}
	\label{fig:sigma-fine-tuning}
\end{figure}

Standard deviation ($\sigma$) plays a crucial role in sentence similarity calculation (Equation \ref{eq:sent_sim}). Therefore, to fix the most suited $\sigma$ value sixty-three different values were experimented on. These values ranged from $10^{-12}$ to $10$ on regular intervals. After experimentation, $5\times10^{-11}$ was fixed as the value for $\sigma$ that gives the most representative semantic relation between sentences. The result of the fine-tuning process is shown in Fig. \ref{fig:sigma-fine-tuning}.
\subsubsection{Different Sentence Extraction Techniques From Clusters: } \label{subsubsec:different-ranking-techniques-inside-clusters}
We have examined two sentence extraction methods to pick the most representative sentence from each cluster. Firstly, the lead extraction method is used to select sentences from the clusters based on their order of appearance in the input document. This ensures a higher proportion of earlier sentences appear in the final summary. Because, generally the earlier sentences in an input contain more information on the context of the input document. Secondly, we examined the TF-IDF ranking technique where we extracted sentences based on their TF-IDF score. The sentence with the highest TF-IDF score in a cluster is selected as the representative sentence. We examined the two methods on our Self-Curated dataset with $0.2$ as a summary proportion. In Table \ref{tab:ranking}, the TF-IDF ranking is shown to perform better than the lead extraction method in the Self-Curated dataset.
\begin{table}[]
	\centering
	\caption{Comparison of Result for different ranking techniques}
	\begin{tabular}{lccc}\hline
		Method      	& Rouge-1       & Rouge-2       & Rouge-LCS     \\\hline
		Lead extraction	& 0.47          & 0.36          & 0.43          \\
		TF-IDF ranking	& \textbf{0.50} & \textbf{0.40} & \textbf{0.46} \\\hline
	\end{tabular}
	\label{tab:ranking}
\end{table}
\subsection{Comparison}\label{subsec:comparison}
The performance of the proposed method (WGSS) is compared with BenSumm \cite{chowdhury-etal-2021-tfidf-clustering}, SASbSC \cite{roychowdhury-etal-2022-spectral-base} and LexRank \cite{Erkan-lexRank-2004} on four datasets (Self-Curated (SC), Kaggle, BNLPC, Github) using the ROUGE metrics. For the comparison step, we fixed the proportion of the summary of WGSS method as $0.2$. The comparative results of this evaluation are shown in Table \ref{tab:result_comparison-1} where our proposed model performs 11.9\%, 24.1\% and 16.2\% better than SASbSC in Rouge-1, Rouge-2 and Rouge-LCS respectively on the self-curated dataset. It performs 68.9\%, 95.4\% and 84.6\% better in Rouge-1, Rouge-2 and Rouge-LCS respectively than BenSumm on the Kaggle dataset. It also performs 3\% and 2.6\% better in Rouge-2 and Rouge-LCS respectively and ties in Rouge-1 with SASbSC using the BNLPC dataset. It performs 58\%, 86.4\%, and 67.9\% better in Rouge-1, Rouge-2 and Rouge-LCS respectively than BenSumm on the GitHub dataset.
\begin{table}[]
	\centering
	\caption{Comparison of average Rouge scores between graph based extractive summarization models on 4 datasets}
	\begin{tabular}{llccc} \hline
		Dataset 		& Model                                                & Rouge-1       & Rouge-2       & Rouge-LCS     \\\hline
		Self-curated	& WGSS (Proposed)                                     & \textbf{0.47} & \textbf{0.36} & \textbf{0.43} \\
		& BenSumm \cite{chowdhury-etal-2021-tfidf-clustering}  & 0.41          & 0.29          & 0.36          \\
		& SASbSC \cite{roychowdhury-etal-2022-spectral-base}   & 0.42          & 0.29          & 0.37          \\
		& LexRank \cite{Erkan-lexRank-2004}                    & 0.22          & 0.14          & 0.20          \\\hline
		Kaggle			& WGSS (Proposed)                                     & \textbf{0.49} & \textbf{0.43} & \textbf{0.48} \\
		& BenSumm \cite{chowdhury-etal-2021-tfidf-clustering}  & 0.29          & 0.22          & 0.26          \\
		& SASbSC \cite{roychowdhury-etal-2022-spectral-base}   & 0.23          & 0.12          & 0.18          \\
		& LexRank \cite{Erkan-lexRank-2004}                    & 0.24          & 0.16          & 0.22          \\\hline
		BNLPC 			& WGSS (Proposed)                                     & \textbf{0.41} & \textbf{0.34} & \textbf{0.40} \\
		& BenSumm \cite{chowdhury-etal-2021-tfidf-clustering}  & 0.36          & 0.28          & 0.34          \\
		& SASbSC \cite{roychowdhury-etal-2022-spectral-base}   & \textbf{0.41} & 0.33          & 0.39          \\
		& LexRank \cite{Erkan-lexRank-2004}                    & 0.26          & 0.19          & 0.24          \\\hline
		Github          & WGSS (Proposed)                                     & \textbf{0.49} & \textbf{0.41} & \textbf{0.47} \\
		& BenSumm \cite{chowdhury-etal-2021-tfidf-clustering}  & 0.31          & 0.22          & 0.28          \\
		& SASbSC \cite{roychowdhury-etal-2022-spectral-base}   & 0.30          & 0.18          & 0.24          \\
		& LexRank \cite{Erkan-lexRank-2004}                    & 0.22          & 0.14          & 0.20          \\\hline
	\end{tabular}
	\label{tab:result_comparison-1}
\end{table}

\begin{figure}[]
	\centering
	\includegraphics[width=5in]{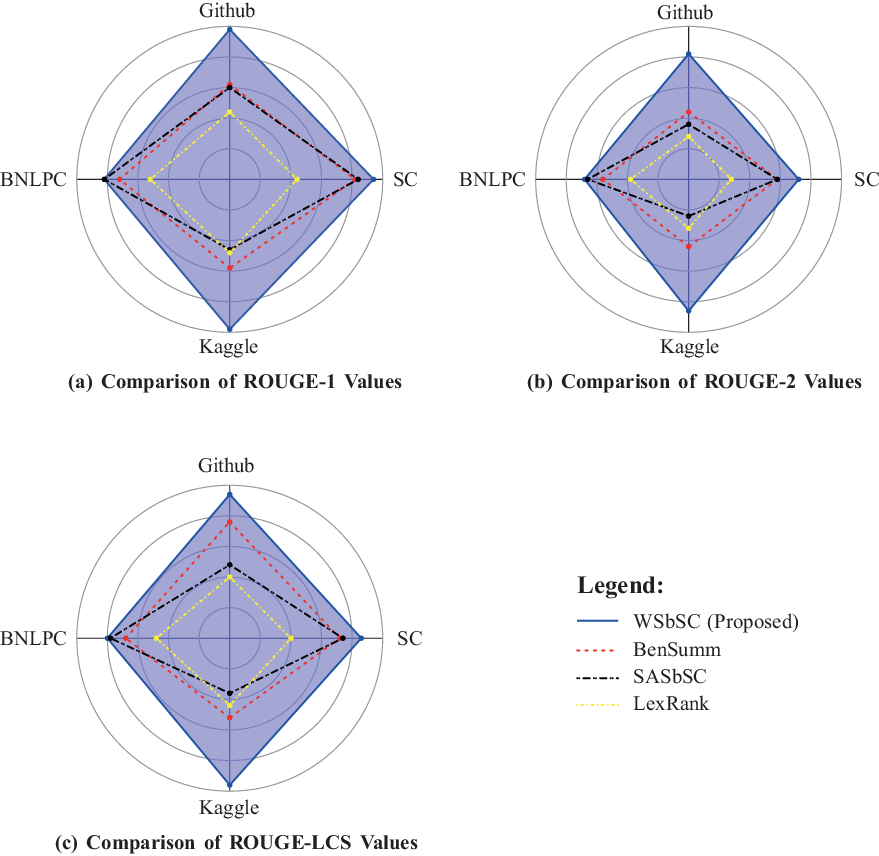}
	\caption{The Radar chart of the models of being compared on four datasets at once}
	\label{fig:radarchart}
\end{figure}

These results are further visualized into three radar charts in Fig. \ref{fig:radarchart} to compare the performance of the models on different Rouge metrics. As stated in the charts, the proposed method performs consistently and uniformly across all the datasets regardless of their quality. But other models, such as BenSumm perform well in three datasets (SC, GitHub, BNLPC) but fail in the Kaggle dataset. Similarly, SASbSC performs well in SC and BNLPC datasets but its performance decreases sharply in Kaggle and GitHub datasets. LexRank although performs consistently across all datasets is far lower on average. %The performance drop of the SASbSC and BenSumm methods in the Github and Kaggle datasets can be attributed to relative low quality of the provided summaries. 

\begin{figure}
	\centering
	\includegraphics[width=5in]{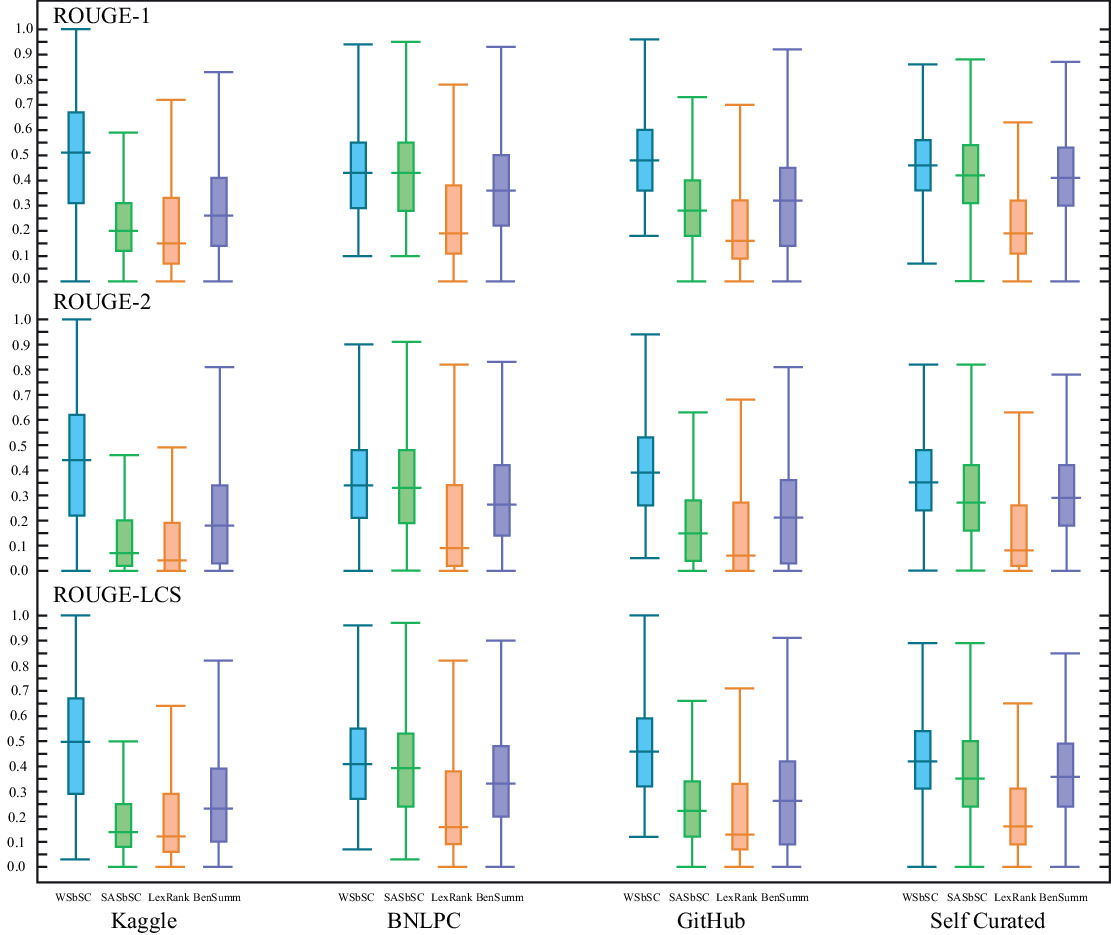}
	\caption{Boxplot chart for performance of the models on four datasets}
	\label{fig:box-rouge}
\end{figure}

The consistency of the proposed method is further visualized using the boxplot charts in Fig. \ref{fig:box-rouge}. It shows, that only SASbSC performs similarly to our model on the BNLPC dataset. In every other case, WGSS shows a smaller interquartile range and a higher median value than the other three models on all datasets. This performance is repeated in Rouge-2 and Rouge-LCS metrics. Thus the result analysis in Table \ref{tab:result_comparison-1} and Fig. \ref{fig:box-rouge} \& \ref{fig:radarchart}, WGSS is the most accurate and reliable Bengali extractive text summarization model.
\subsection{Implementation Into Other Languages}\label{subsec:implementation-into-other-languages}
The proposed model is language-independent thus, it can be extended into other languages too. For this, only a language-specific tokenizer, a stop-word list and a word embedding dataset are required. We implemented this model on three non-Bengali datasets to show the language-independent nature of the model. To evaluate the quality of the generated summaries, we tried to find evaluation datasets for summarization on other low-resource languages. But could only find relevant datasets in three other languages i.e. Hindi, Marathi and Turkish. We adopted the proposed model into these three low-resource languages to check how well it performs.
\begin{table}[]
	\centering
	\caption{Comparison of Result of proposed summarization method in other low-resource languages}
	\begin{tabular}{lccc}\hline
		Language              	& Rouge-1   & Rouge-2   & Rouge-LCS \\\hline
		Bengali (Self-curated)	& 0.47      & 0.36      & 0.43      \\
		Bengali (Kaggle)   		& 0.49      & 0.43      & 0.48      \\
		Bengali (BNLPC)   		& 0.41      & 0.34      & 0.40      \\
		Bengali (Github)   		& 0.49      & 0.41      & 0.47      \\
		Bengali (Average)       & 0.47      & 0.38      & 0.44      \\\hline
		Hindi                   & 0.40      & 0.26      & 0.36      \\\hline
		Marathi                 & 0.50	    & 0.42      & 0.50      \\\hline
		Turkish                 & 0.48      & 0.39      & 0.47      \\\hline
	\end{tabular}
	\label{tab:other_language}
\end{table}
Table \ref{tab:other_language} shows the result of the proposed WGSS method for extractive summarization in other low-resource languages. In this table, the results in Marathi and Turkish are slightly better than the results in Bengali. Although it performs slightly lower in Hindi, the score is still similar to Bengali. To evaluate the model's performance on Hindi, we used a Kaggle dataset\footnote{\textit{\href{https://www.kaggle.com/datasets/disisbig/hindi-text-short-and-large-summarization-corpus/}{https://www.kaggle.com/datasets/disisbig/hindi-text-short-and-large-summarization-corpus/}}} produced by Gaurav Arora. For the Marathi, we used another Kaggle dataset\footnote{\textit{\href{https://www.kaggle.com/datasets/ketki19/marathi}{https://www.kaggle.com/datasets/ketki19/marathi}}} produced by Ketki Nirantar. For the Turkish language, we used a GitHub dataset\footnote{\textit{\href{https://www.github.com/xtinge/turkish-extractive-summarization-dataset/blob/main/dataset/XTINGE-SUM_TR_EXT/xtinge-sum_tr_ext.json}{https://wwww.github.com/xtinge/turkish-extractive-summarization-dataset}}} produced by the XTINGE \cite{Demir-2024-xtinge_turkish_extractive} team for evaluation. 
\section{Discussion}\label{sec:discussion}
The results presented in Table \ref{tab:result_comparison-1} \& \ref{tab:other_language} and in Fig. \ref{fig:radarchart} \& \ref{fig:box-rouge} highlight the effectiveness of the proposed WGSS model for extractive text summarization in Bengali, as well as its adaptability to other low-resource languages. This section analyses the comparative results, the strengths and limitations of the proposed method, and potential areas for further research.

As evidenced by the results shown in tables and figures in comparison section \ref{subsec:comparison}, the WGSS model consistently outperforms other graph-based extractive text summarization models, namely BenSumm \cite{das-2022-tfidf}, LexRank \cite{Erkan-lexRank-2004}, and SASbSC \cite{roychowdhury-etal-2022-spectral-base}. The proposed model shows better performance compared to the other three methods on Rouge-1, Rouge-2, and Rouge-LCS metrics. This performance improvement is due to the novel approach to calculating sentence similarity. While calculating sentence similarity, taking the geometric mean of individual similarity between word pairs overcomes the lack of local word correspondence faced by the averaging vector method \cite{roychowdhury-etal-2022-spectral-base}. The Gaussian kernel-based word similarity provides a precise semantic relationship between sentences which further contributes to the performance improvement. Another reason for performance improvement is the usage of spectral clustering which is very effective in capturing irregular cluster shapes.

Our proposed strategy for calculating sentence similarity is more suited than existing methods for comparing two sets of vectors in the context of language such as Earth Movers Distance (EMD) \cite{Rubner-19998-emd}, Hausdorff Distance \cite{hausdorff-1914-hausdorff-distance}, Procrustes Analysis \cite{Gower-1975-procrustes-distance}. Whereas, EMD and Procrustes Analysis are computationally expensive and irrelevant for word vectors due to involving scaling or rotating vectors which do not hold any semantic meaning. Another method, Hausdorff distance \cite{hausdorff-1914-hausdorff-distance} calculates the highest possible distance between vectors from two sets. Although not as expensive as EMD and Procrustes Analysis, it is easily influenced by outlier words due to only considering the worst-case scenario.

On the other hand, the proposed method focuses on local vector similarity between two sets which is more important for words. The Gaussian similarity function captures the proximity of points smoothly, providing a continuous value for similarity between two words in a normalized way. Gaussian similarity is also robust against small outliers due to being a soft similarity measure. Taking the geometric mean of Gaussian similarities also helps to smooth over similarity values for outlier words.

One of the key strengths of this proposed method is the reduction of redundancy in the output summary which is a common challenge in extractive summarization methods. This is achieved by grouping semantically similar sentences together. The use of spectral clustering for the grouping task improves the performance by allowing flexibility for cluster shapes. Another key strength of WGSS is the improved sentence similarity calculation technique over the word averaging method used by SASbSC \cite{roychowdhury-etal-2022-spectral-base}, which dilutes the semantic meaning of a sentence. The scalability of our method across languages is another advantage due to it requiring very few language-specific resources. This scalability is demonstrated in the experiments with Hindi, Marathi, and Turkish languages (Table \ref{tab:other_language}).

Despite its advantages, the WGSS model does face some challenges. The model heavily relies on pre-trained word embeddings, which may not always capture the full details of certain domains or newly coined terms. The FastText \cite{grave-etal-2018-fasttext} dataset used here is heavily reliant on Wikipedia for training which could introduce some small unforeseen biases. The model also does not take into account the order in which words appear in a sentence or when they form special noun or verb groups. 

The proposed WGSS model represents a significant advancement in Bengali extractive text summarization due to its ability to accurately capture semantic similarity, reduce redundancy, maximize coverage and generalize across languages. The results of this study demonstrate the robustness and adaptability of the WGSS model, offering a promising direction for future research in multilingual extractive summarization.

\section{Conclusion}\label{sec:conclusion}
In this study, we proposed the WGSS method for Bengali extractive text summarization which is also extended to other low-resource languages. The process uses the geometric mean of Gaussian similarities between individual word pairs to identify deeper semantic relationships within two sentences. This sentence similarity calculation method shows better text summarization performance than recent graph-based techniques. Using the sentence similarity, an affinity matrix is built to be clustered into groups. We extract one sentence from each cluster to generate the summaries. This approach improves the coherence and relevance of the generated summaries by minimizing redundancy and maximizing topic coverage. 43.2\% average improvement across three ROUGE metrics proves the versatility and robustness of the proposed method. WGSS can also be extended into other languages as shown through the results in Hindi, Marathi and Turkish languages. 
%This proves the generalizability of the method. 
It addresses the need for an effective summarization technique in the Bengali language.

This work contributes to the growing body of computational linguistics research focused on low-resource languages like Bengali. The results showed the strengths of the proposed WGSS method compared to several baseline techniques. Despite the promising results, WGSS may face limitations on highly specialized or domain-specific texts where deeper linguistic features beyond word similarity such as word order, self-attention, sophisticated post-processing techniques etc. could be considered in the future.

\section{Acknowledgement}\label{sec:acknowledgement}
This research was supported partially by the Bangladesh Research and Education Network (BdREN) cloud computing resources. An expert linguistic team also helped us to build the evaluation dataset. This team consisted of Arnab Das Joy, Noshin Tabassum Dina, Kazi Farhana Faruque, Md. Morshaline Mojumder, Mrinmoy Poit, Md. Mahfujur Rahman, Effaz Rayhan, Asaduzzaman Rifat, Md. Rohan Rifat, Musfiqur Rahman Shishir and Akidul Islam Jim. All of them are undergrad students of the Institute of Information Technology, University of Dhaka.

\bibliographystyle{acm}  
\bibliography{references}

\end{document}